\documentclass{article}

\usepackage{ifthen}
\newboolean{neurips_version}
\setboolean{neurips_version}{true}
\ifthenelse{\boolean{neurips_version}}
{\PassOptionsToPackage{numbers, compress}{natbib}
 \usepackage[preprint]{neurips_2022}
}
{
\usepackage{PRIMEarxiv}
 \usepackage[numbers]{natbib}
}

\usepackage[utf8]{inputenc} 
\usepackage[T1]{fontenc}    
\usepackage{hyperref}       
\usepackage{url}            
\usepackage{booktabs}       
\usepackage{amsfonts,amsmath,amsthm,amssymb}       
\usepackage{nicefrac}       
\usepackage{microtype}      
\usepackage{xcolor}         
\usepackage[disable]{todonotes}
\usepackage{enumitem}
\usepackage{caption}

\usepackage{mathtools} 
\usepackage{tikz}
\usetikzlibrary{calc}
\usepackage{pgfplots}
\usepackage{graphicx}

\usepackage{amsmath}
\usepackage{algorithm, algpseudocode}
\usepackage{amsfonts}  

\usepackage[english]{babel}
\usepackage{amsthm}

\usepackage{subfig}
\usepackage[export]{adjustbox}

\usepackage{hhline}
\usepackage{makecell, caption, booktabs}
\usepackage{siunitx}

\usepackage{multirow}

\usepackage{amsmath, nccmath}
\usepackage{bigstrut}

\usepackage{booktabs}

\usepackage{lipsum}
\graphicspath{{media/}}     

\usepackage{CJKutf8}
\usepackage[framed,numbered,autolinebreaks,useliterate]{mcode}

\title{ Multinomial Logistic Regression Algorithms \\ via  \\ Quadratic Gradient}

\author{ \href{https://orcid.org/0000-0003-0378-0607}{\includegraphics[scale=0.06]{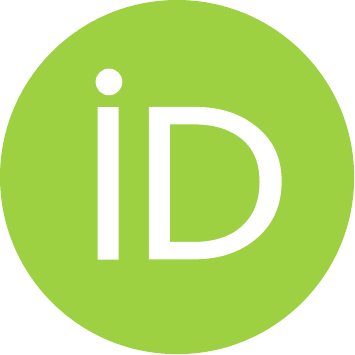}\hspace{1mm}John Chiang} \\                             
                                      \\
	\texttt{john.chiang.smith@gmail.com} 
}

\date{}

\theoremstyle{remark}

\renewcommand{\epsilon}{\varepsilon}

\makeatletter
\def\namedlabel#1#2{\begingroup
    #2%
    \def\@currentlabel{#2}%
    \phantomsection\label{#1}\endgroup
}
\makeatother

\algnewcommand{\LeftComment}[1]{\Statex \(\triangleright\) #1}
\algnewcommand{\LineCommentStep}[1]{\Statex \textbf{[Step #1]:} }
\makeatletter
\newlength{\trianglerightwidth}
\algnewcommand{\LineComment}[1]{\Statex \hskip\ALG@thistlm $\triangleright$ #1}
\algnewcommand{\LineCommentCont}[1]{\Statex \hskip\ALG@thistlm%
  \parbox[t]{\dimexpr\linewidth-\ALG@thistlm}{\hangindent=\trianglerightwidth \hangafter=1 \strut$\triangleright$ #1\strut}}
\algnewcommand{\LeftLineCommentCont}[1]{\Statex \hskip\ALG@thistlm%
  \parbox[t]{\dimexpr\linewidth-\ALG@thistlm}{\leftskip=\algorithmicindent \hangindent=\trianglerightwidth \hangafter=1 \strut$\triangleright$ #1\strut}}

\newcommand{\mysplit}[1]{%
  \begin{tabular}{@{}c@{}}
    #1
  \end{tabular}
  }
  
\begin{document}

\maketitle

\begin{abstract}%
Multinomial logistic regression, also known by other names such as multiclass logistic regression and softmax regression, is a fundamental classification  method that generalizes binary logistic regression to multiclass problems.  A recently work~\cite{chiang2022privacy} proposed a faster gradient called $\texttt{quadratic gradient}$  that can accelerate the binary logistic regression training, and presented an enhanced  Nesterov's accelerated gradient (NAG) method for binary logistic regression. 
In this paper, we extend this work to multiclass logistic regression and propose an enhanced Adaptive Gradient Algorithm (Adagrad) that can accelerate the original Adagrad method. We test the enhanced NAG method and the enhanced Adagrad method on some multiclass-problem datasets. Experimental results show that both enhanced methods converge faster than their original ones respectively. 
\todo{Show precise somewhere that our results are new in the iid setting as well.}

\end{abstract}

\listoftodos

\section{Introduction}

Logistic regression (LR) is a classical model in machine learning applied for estimating conditional probabilities. This model has been widely used in a variety of specific domains such as health and finance, both for binary classification and multi-class problems. Although the theory for binary logistic regression is well developed~\citep{SmartandVercauteren_SIMD}, its multiclass extension is much less studied. 

Multinomial logistic regression is a decision-making task where examples consist of a feature vector and a categorical class label with more than two choices. For each input of a  feature vector representing the example to be classified, the multinomial logistic regression model is expected to output a  prediction of probabilities of the label of the example. The model performance can be measured by their log-loss and accuracy. 

Motivated by its widely used in real-world applications, effects have been made to find efficient methods for multiclass LR.  B\"ohning ~\citep{bohning1992multinomial} proposed a fixed good lower bound for multiclass LR. Our work 
 is most related to Chiang~\citep{chiang2022privacy}
 and can be seen as the following work to \citep{chiang2022privacy}. Chiang~\citep{chiang2022privacy} proposed a faster gradient variant named $\texttt{quadratic gradient}$ and presented an enhanced NAG method for binary LR with the help of $\texttt{quadratic gradient}$. This work extends binary LR to multiclass LR problems and presents the enhanced Adagrad method for the multiclass LR. 

\section{Preliminaries}
Chiang~\citep{chiang2022privacy} proposed a faster gradient variant named $\texttt{quadratic gradient}$ and claimed that quadratic gradient can unite the first-order gradient method and the second-order Newton's method. This faster gradient variant can be seen as an extension of the simplified fixed Hessian~\citep{bonte2018privacy} and be built by constructing a diagonal substituted matrix that suffices the convergence condition of the fixed Hessian method~\citep{bohning1988monotonicity}.

\paragraph{Chiang's Quadratic Gradient} Supposing that  a differentiable scalar-valued function $F(\mathbf x)$ has its gradient $g$ and Hessian matrix $H$. To maximize the function $F(\mathbf x)$, we try to find a  good lower bound matrix $\bar H \le H$; To minimize the function $F(\mathbf x)$, we need to find a good upper bound $\bar H$ such that $H \le \bar H$. Here ``$ \le $'' is in the Loewner order and $A \le B$ means that $A - B$ is non-negative. We first build a diagonal matrix $\bar B$ from the good bound matrix $\bar H$  as follows: 

 \begin{equation*}
  \begin{aligned}
   \bar B = 
\left[ \begin{array}{cccc}
  \frac{1}{ \epsilon + \sum_{i=0}^{d} | \bar h_{0i} | }   & 0  &  \ldots  & 0  \\
 0  &   \frac{1}{ \epsilon + \sum_{i=0}^{d} | \bar h_{1i} | }  &  \ldots  & 0  \\
 \vdots  & \vdots                & \ddots  & \vdots     \\
 0  &  0  &  \ldots  &   \frac{1}{ \epsilon + \sum_{i=0}^{d} | \bar h_{di} | }  \\
 \end{array}
 \right], 
   \end{aligned}
\end{equation*}
where $\bar h_{ji}$ is the elements of the matrix $\bar H$ and $\epsilon $  an arbitrarily small positive number.

The quadratic gradient $G$ for function $F(\mathbf x)$ is defined as  $G = \bar B \cdot g$  and have the same dimension as the naive gradient $g$. The way to use it is much the same as the first-order gradient method, except that we need to replace the naive gradient with the quadratic gradient and increase the learning rate by $1$. For example, $\mathbf x = \mathbf x + \eta G$ and $\mathbf x = \mathbf x - \eta G$ are to maximise the function $F(\mathbf x)$ and minimise the function $F(\mathbf x)$ respectively. Note that the new learning rate $\eta$ should be no less than $1$ for efficiency and convergence speed. Moreover, sophisticated gradient (descent) methods such as NAG and Adagrad can be applied in practice to improve the performance further. 

Bonte and Vercauteren~\cite{bonte2018privacy} presented the Simplified Fixed Hessian (SFH) method for binary LR by constructing a simplified diagonal fixed Hessian from the good lower bound matrix proposed by B\"ohning and Lindsay \cite{bohning1988monotonicity}. In real-world applications, it is not easy to find the fixed Hessian matrix, and sometimes it even does not exist or is not a good lower bound of the Hessian matrix. In these cases, we can build the quadratic gradient directly from the Hessian matrix itself, rather than struggling to find a fixed Hessian substitute. The quadratic gradient doesn't stick to a ``fixed''  replaced matrix but only adheres to a good bound matrix. Letting the Hessian matrix $H$ itself be the good  bound $\bar H$, we can see that the diagonal matrix $\bar B$ built directly from $H$  suffices the theory of the fixed Hessian method.

For maxisizing $F(  \mathbf  x )$, we need to find a matrix $\bar H$ such that $\bar H \le H$ to follow the (simplified) fixed Hessian method. Chiang~\citep{chiang2022privacy} has already proved that the above given $\bar B$ is a lower bound to $\bar H$: $\bar B \le \bar H \le  H$.
 
The problem of minimising  $F(\mathbf x)$, however, can be seen as to  maxisize the function $G(\mathbf x) = -F(\mathbf x)$. Function $G(\mathbf x)$ has its gradient
$- g$
 and Hessian matrix $ - H$. To maxisize $G(\mathbf x)$, we need to find a matrix $\bar H$ such that $\bar H \le -H$ and get the iteration: $\mathbf x = \mathbf x + (\bar H)^{-1} (-g)$ or $\mathbf x = \mathbf x - (\bar  H)^{-1} g$. 
 Namely, we need to find the lower bound $\bar  H $ of $-H$ or the upper bound $- \bar  H$ of $H$ such that $- \bar  H \ge H$ and then build the quadratic gradient from $\bar  H$ or $- \bar  H$. The constructions of quadratic gradient for $\bar H $ and $- \bar H $ result in the same quadratic gradient.

In conclusion, to calculate the quadratic gradient, we can find a matrix $\bar H \le H$ for the maximization problem or a matrix $\bar  H \ge  H$ for the  minimization problem. A better way is to construct the $\bar B$ from the Hessian matrix $H$ directly and then try to find a good upper bound for each diagonal element of $\bar B$ if there exists one.

\section{Technical details}
%
Given an  dataset matrix $X \in \mathbb{R}^{n \times (1 + d)}$, each row of which represents a record with $d$ features plus the first element $1$. The outcome  $Y \in \mathbb{N}^{n \times 1}$ to 
  the input dataset $X$ conssits of corresponding  $n$  class labels. Let each record $x_{[i]}$ has an observation result $y_{[i]}$ with $c$ possible classes labelled as $y_{i} \in \{ 0, 1, \cdots, c-1 \}$. The multiclass logistic regression model has $c$ paremeter vectors $w_{i}$ of size $ (1 + d)$, which  form  the parameter matrix $W$.  Each vector $\mathbf w_{i}$ is used to model the probability of each class label. 

\begin{align*}
  X &= 
 \begin{bmatrix}
 \textsl x_{1}      \\
 \textsl x_{2}      \\
 \vdots          \\
 \textsl x_{n}      \\
 \end{bmatrix}
 = 
 \begin{bmatrix}
 x_{[1][0]}    &   x_{[1][1]}   &  \cdots  & x_{[1][d]}   \\
 x_{[2][0]}    &   x_{[2][1]}   &  \cdots  & x_{[2][d]}   \\
 \vdots    &   \vdots   &  \ddots  & \vdots   \\
 x_{[n][0]}    &   x_{[n][1]}   &  \cdots  & x_{[n][d]}   \\
 \end{bmatrix} 
 ,   
 Y =  
 \begin{bmatrix}
 y_{1}     \\
 y_{2}     \\
 \vdots         \\
 y_{n}     \\
 \end{bmatrix},  \\
 W &=  
 \begin{bmatrix}
 \textsl w_{0}     \\
 \textsl w_{1}     \\
 \vdots         \\
 \textsl w_{c-1}     \\
 \end{bmatrix}  =
 \begin{bmatrix}
 w_{[0][0]}    &   w_{[0][1]}   &  \cdots  & w_{[0][d]}   \\
 w_{[1][0]}    &   w_{[1][1]}   &  \cdots  & w_{[1][d]}   \\
 \vdots    &   \vdots   &  \ddots  & \vdots   \\
 w_{[c-1][0]}    &   w_{[c-1][1]}   &  \cdots  & w_{[c-1][d]}   \\
 \end{bmatrix}. 
\end{align*} 
\begin{align*} 
 \pi &= [w_{0}, w_{1}, \hdots , w_{c} ]^T      \\
     &= [ w_{[0][0]},  w_{[0][1]},        \cdots ,  w_{[0][d]}, w_{[1][0]},w_{[1][1]},     \cdots,   w_{[1][d]}, \cdots, w_{[c-1][0]},w_{[c-1][1]},       \cdots,   w_{[c-1][d]} ]^T.  \\
\end{align*} 

MLR aims to find the best parameter matrix $W$ such that the algorithm can model the probabilities of each class label for each record by the softmax function and output the class label with the maximum probability.
  
\begin{align*}
 P &=  
 \begin{bmatrix}
 p_{[1][0]}    &   p_{[1][1]}   &  \cdots  & p_{[1][c-1]}   \\
 p_{[2][0]}    &   p_{[2][1]}   &  \cdots  & p_{[2][c-1]}   \\
 \vdots    &   \vdots   &  \ddots  & \vdots   \\
 p_{[n][0]}    &   p_{[n][1]}   &  \cdots  & p_{[n][c-1]}   \\
 \end{bmatrix}  \\
   &=  
 \begin{bmatrix}
 Prob(y=0|\textsl  x = \textsl x_{1})   &   Prob(y=1|\textsl  x = \textsl  x_{1})   &  \cdots  & Prob(y=c-1|\textsl  x = \textsl x_{1})   \\
 Prob(y=0| \textsl x = \textsl  x_{2})    &   Prob(y=1| \textsl x = \textsl x_{2})   &  \cdots  & Prob(y=c-1| \textsl x = \textsl  x_{2})   \\
 \vdots    &   \vdots   &  \ddots  & \vdots   \\
 Prob(y=0|\textsl  x = \textsl x_{n})    &   Prob(y=1| \textsl  x = \textsl x_{n})   &  \cdots  & Prob(y=c-1|\textsl  x = \textsl x_{n})   \\
 \end{bmatrix},  \\
\end{align*} 
where $$ p_{[i][j]} = Prob(y=j|\textsl x= \textsl  x_{i}) = \frac{ \exp (\textsl  x_{i}\cdot \textsl{w}_{j}^T ) }{ \sum_{k=0}^{c-1} \exp (\textsl  x_{i}\cdot\textsl{w}_{k}^T ) } $$ and $$\sum_{j=0}^{c-1} p_{[i][j]} = Prob(y=0|\textsl x= \textsl  x_{i}) + \dots + Prob(y=c-1|\textsl x= \textsl x_{i}) = 1 \ \text{for}\  1 \le i \le n. $$

Multiclass LR is to maxsize $$L = \prod _{i=1}^n Prob(y=y_{i}|\textsl x=\textsl x_{i}) = \prod _{i=1}^n \frac{  \exp (\textsl x_{i}\cdot\textsl{w}_{y_{i}}^T ) }{ \sum_{k=0}^{c-1} \exp (\textsl x_{i}\cdot\textsl{w}_{k}^T ) }$$ or its log-likelihood function $$\ln L = \sum_{i=1}^n  [ \textsl x_{i}\cdot\textsl{w}_{y_{i}}^T  - \ln  \sum_{k=0}^{c-1} \exp (\textsl x_{i}\cdot\textsl{w}_{k}^T) ]. $$

For simplicity in presentation, we first consider the case of dataset $X$ consisting of just one sample $\textsl{x}_{1}$ with the label class $y_{1}$. 

\begin{align}
\frac{ \partial \ln L }{  \partial \textsl{w}_{k}  }_{(k = y_{1})} =\textsl x_{1} \cdot 1 - \frac{\textsl x_{1} \cdot \exp (\textsl x_{1}\cdot\textsl{w}_{y_{1}}^T ) }{ \sum_{k=0}^{c-1} \exp (\textsl x_{1}\cdot\textsl{w}_{k}^T ) }   =\textsl x_{1} \cdot [1 - Prob(y=y_{1}|\textsl x=\textsl x_{1}) ] \label{FF} \\
 \frac{ \partial \ln L }{  \partial \textsl{w}_{k}  }_{(k \ne y_{1})} =\textsl x_{1} \cdot 0 - \frac{\textsl x_{1} \cdot \exp (\textsl x_{1}\cdot\textsl{w}_{y_{1}}^T ) }{ \sum_{k=0}^{c-1} \exp (\textsl x_{1}\cdot\textsl{w}_{k}^T ) }   =\textsl x_{1} \cdot [0 - Prob(y=y_{1}|\textsl x= \textsl x_{1}) ] \label{GG}
\end{align} 
Here comes the one-hot encoding method with its practical use in uniting formulas $(\ref{FF})$ and $(\ref{GG})$.

One-hot encoding is the most widely adopted technique of representing categorical data  as a binary vecotr. For a class label $y_i$ with $c$ possible values, one-hot encoding converts $y_i$ into a vector $\bar{ \textsl y_i }$ of size $c$ such that $\bar{\textsl y_i}[j] = 1$ if $j = y_i$ and $\bar{\textsl y_i}[j] = 0$ otherwise:
$y_{i} \xmapsto{ \text{one-hot encoding} } \bar{ \textsl y_{1} }  = [ y_{[1][0]},  y_{[1][1]}, \hdots,  y_{[1][c-1]}]$. Therefore,  the outcome  $Y \in \mathbb{N}^{n \times 1}$ can be transformed by this encoding method into $\bar Y \in \mathbb{N}^{n \times c}:$ 

\begin{align*}
 Y =  
 \begin{bmatrix}
 y_{1}     \\
 y_{2}     \\
 \vdots         \\
 y_{n}     \\
 \end{bmatrix} 
\xmapsto{ \text{one-hot encoding} } 
\bar Y =
 \begin{bmatrix}
 \bar{ \textsl{y}_{1} }     \\
 \bar{ \textsl{y}_{2} }     \\
 \vdots         \\
 \bar{ \textsl{y}_{n} }     \\
 \end{bmatrix} 
 =
 \begin{bmatrix}
 y_{[1][1]}    &   y_{[1][2]}   &  \cdots  & y_{[1][c-1]}   \\
 y_{[2][1]}    &   y_{[2][2]}   &  \cdots  & y_{[2][c-1]}   \\
 \vdots    &   \vdots   &  \ddots  & \vdots   \\
 y_{[n][1]}    &   y_{[n][2]}   &  \cdots  & y_{[n][c-1]}   \\
 \end{bmatrix}. 
\end{align*} 
We can hence obtain a single unified formula with the help of the one-hot encoding of $y_{1}$:
$$ \frac{ \partial \ln L }{  \partial \textsl{w}_{k}  } =  y_{[1][k]} \cdot \textsl x_{1} - \frac{\textsl x_{1} \cdot \exp (\textsl x_{1}\cdot\textsl{w}_{y_{1}}^T ) }{ \sum_{k=0}^{c-1} \exp (\textsl x_{1}\cdot\textsl{w}_{k}^T ) }   = \textsl x_{1} \cdot [ y_{[1][k]} - Prob(y=y_{1}|\textsl x=\textsl x_{1}) ], $$

which yields the gradient of $\ln L$:

\begin{align*}
 \nabla = \frac{ \partial \ln L }{  \partial \pi  } &= \Big [ \frac{ \partial \ln L }{  \partial \textsl{w}_{[0]} }, \frac{ \partial \ln L }{ \partial \textsl{w}_{[1]} } , \hdots, \frac{ \partial \ln L }{  \partial \textsl{w}_{[c-1]} }  \Big ]^T  \\  
&= 
\Big [ \frac{ \partial \ln L }{  \partial {w}_{[0][0]} },\hdots,\frac{ \partial \ln L }{  \partial {w}_{[0][d]} },  \frac{ \partial \ln L }{  \partial {w}_{[1][0]} },\hdots,\frac{ \partial \ln L }{  \partial {w}_{[1][d]} }, \hdots, \frac{ \partial \ln L }{  \partial {w}_{[c-1][0]} },\hdots,\frac{ \partial \ln L }{  \partial {w}_{[c-1][d]} } \Big ]^T  \\
&= \Big [ \mathbf{x}_1 \cdot [\bar y_{[1][0]} - p_{[1][y_1]}], \mathbf{x}_1 \cdot [\bar y_{[1][1]} - p_{[1][y_1]} ], \cdots, \mathbf{x}_1 \cdot [\bar y_{[1][c-1]} - p_{[1][y_1]} ]  \Big ]
\end{align*}
or could be reshaped to the same dimension as  $W$:
$$ [ \nabla ] = [ \bar y_1 - p_1 ]^T \cdot x_1 $$ for the convenience in gradient computation: $W = W +  [ \nabla ] $.

The log-likelihood $\ln L$ should be seen as a multivariate function  of $[(1+c) (1+d)]$ variables,  whose gradient is a column vector of size $[(1+c)(1+d)]$ and whose Hessian matrix $\nabla^2$ is literally  a square matrix of order $[(1+c)(1+d)].$

\begin{align*}
 \nabla^2 &=  
 \begin{bmatrix}
 \frac{ \partial^2 \ln L }{  \partial \textsl{w}_{[0]} \partial \textsl{w}_{[0]} }    &   \frac{ \partial^2 \ln L }{  \partial \textsl{w}_{[0]} \partial \textsl{w}_{[1]} }   &  \cdots  & \frac{ \partial^2 \ln L }{  \partial \textsl{w}_{[0]} \partial \textsl{w}_{[c-1]} }   \\
 \frac{ \partial^2 \ln L }{  \partial \textsl{w}_{[1]} \partial \textsl{w}_{[0]} }    &   \frac{ \partial^2 \ln L }{  \partial \textsl{w}_{[1]} \partial \textsl{w}_{[1]} }   &  \cdots  & \frac{ \partial^2 \ln L }{  \partial \textsl{w}_{[1]} \partial \textsl{w}_{[c-1]} }   \\
 \vdots    &   \vdots   &  \ddots  & \vdots   \\
 \frac{ \partial^2 \ln L }{  \partial \textsl{w}_{[c-1]} \partial \textsl{w}_{[0]} }    &   \frac{ \partial^2 \ln L }{  \partial \textsl{w}_{[c-1]} \partial \textsl{w}_{[1]} }   &  \cdots  & \frac{ \partial^2 \ln L }{  \partial \textsl{w}_{[c-1]} \partial \textsl{w}_{[c-1]} }   \\
 \end{bmatrix}  \\
   &=  
 \begin{bmatrix}
 p_{[1][0]}(p_{[1][0]} - 1) \cdot {\textsl{x}_1}^T\textsl{x}_1   &    p_{[1][0]}p_{[1][1]} \cdot {\textsl{x}_1}^T\textsl{x}_1   &  \cdots  &  p_{[1][0]}p_{[1][c-1]} \cdot {\textsl{x}_1}^T\textsl{x}_1   \\
 p_{[1][1]}p_{[1][0]} \cdot {\textsl{x}_1}^T\textsl{x}_1    &   p_{[1][1]}(p_{[1][1]} - 1) \cdot {\textsl{x}_1}^T\textsl{x}_1    &  \cdots  &  p_{[1][1]}p_{[1][c-1]} \cdot {\textsl{x}_1}^T\textsl{x}_1   \\
 \vdots    &   \vdots   &  \ddots  & \vdots   \\
 p_{[1][c-1]}p_{[1][0]} \cdot {\textsl{x}_1}^T\textsl{x}_1    &    p_{[1][c-1]}p_{[1][1]} \cdot {\textsl{x}_1}^T\textsl{x}_1   &  \cdots  & p_{[1][c-1]}(p_{[1][c-1]} - 1) \cdot {\textsl{x}_1}^T\textsl{x}_1    \\
 \end{bmatrix}  \\
 &=
 \begin{bmatrix}
 p_{[1][0]}(p_{[1][0]} - 1)   &    p_{[1][0]}p_{[1][1]}   &  \cdots  &  p_{[1][0]}p_{[1][c-1]}   \\
 p_{[1][1]}p_{[1][0]}     &   p_{[1][1]}(p_{[1][1]} - 1)     &  \cdots  &  p_{[1][1]}p_{[1][c-1]}   \\
 \vdots    &   \vdots   &  \ddots  & \vdots   \\
 p_{[1][c-1]}p_{[1][0]}     &    p_{[1][c-1]}p_{[1][1]}    &  \cdots  & p_{[1][c-1]}(p_{[1][c-1]} - 1)   \\
 \end{bmatrix}  
 \otimes
   [ {\textsl{x}_1}^T\textsl{x}_1 ],
\end{align*} 
where $\frac{ \partial^2 \ln L }{  \partial \textsl{w}_{[i]} \partial \textsl{w}_{[j]} }  = \frac{ \partial^2 \ln L }{  \partial \textsl{w}_{[j]} \partial \textsl{w}_{[i]} }  =   [p_{[1][i]}(p_{[1][j]} - 1 )] \cdot {\textsl{x}_1}^T\textsl{x}_1$ for $i = j$ and $\frac{ \partial^2 \ln L }{  \partial \textsl{w}_{[i]} \partial \textsl{w}_{[j]} }  = \frac{ \partial^2 \ln L }{  \partial \textsl{w}_{[j]} \partial \textsl{w}_{[i]} }  = [p_{[1][i]}(p_{[1][j]} - 0 )] \cdot {\textsl{x}_1}^T\textsl{x}_1$ for $i \ne j$ and ``$\otimes$'' is the kronecker product.

Kronecker product, unlike the  usual matrix multiplication, is an operation that is perfromed on two matrices of arbitrary size to result in a block matrix. For an $m \times n$ matrix $A$ and a $p \times q$ matrix $B$, the kronecker product of $A$ and $B$ is a $pm \times qn$ block matrix, denoted by $A \otimes B$:
\begin{align*}
 A \otimes B &=  
 \begin{bmatrix}
 a_{11}B    &   a_{12}B   &  \cdots  & a_{1n}B   \\
 a_{21}B    &   a_{22}B   &  \cdots  & a_{2n}B   \\
 \vdots    &   \vdots   &  \ddots  & \vdots   \\
 a_{m1}B    &   a_{m2}B   &  \cdots  & a_{mn}B   \\
 \end{bmatrix}  \\
   &=  
 \begin{bmatrix}
 a_{11}b_{11} & \cdots & a_{11}b_{1q}     &    &    & a_{1n}b_{11} & \cdots & a_{1n}b_{1q}    \\
 \vdots      & \ddots & \vdots          &  \cdots  &  \cdots  & \vdots      & \ddots & \vdots   \\
 a_{11}b_{p1} & \cdots & a_{11}b_{pq}     &    &    & a_{1n}b_{p1} & \cdots & a_{1n}b_{pq}   \\
  & \vdots &      &  \ddots  &          &  &\vdots&   \\
 & \vdots &       &          &  \ddots  &  &\vdots&   \\
 a_{m1}b_{11} & \cdots & a_{m1}b_{1q}     &    &    & a_{mn}b_{11} & \cdots & a_{mn}b_{1q}    \\
 \vdots      & \ddots & \vdots          &  \cdots  &  \cdots  & \vdots      & \ddots & \vdots   \\
 a_{m1}b_{p1} & \cdots & a_{m1}b_{pq}     &    &    & a_{mn}b_{p1} & \cdots & a_{mn}b_{pq}   \\
 \end{bmatrix}  
\end{align*} 
where $a_{ij}$ and $b_{kl}$ are the elements of $A$ and $B$ respectively.

The inverse of the kronecker product $A \otimes B$  exsits if and only if $A$ and $B$ are  both invertable, and can be obtained by 
  \begin{align}
   (A \otimes B)^{-1} = A^{-1} \otimes B^{-1}. \label{KP invertible product property} 
  \end{align} 
This invertible product property can be used to facilitate the calculation of the inverse $\Big [ \frac{ \partial^2 \ln L }{  \partial \pi \partial \pi }  \Big ]^{-1} $ of Hessian matrix ${\nabla^2}$:
\begin{align*}
 \Big [\frac{ \partial^2 \ln L }{  \partial \pi \partial \pi } \Big ]^{-1} =  
   \begin{bmatrix}
 p_{[1][0]}(p_{[1][0]} - 1)   &    p_{[1][0]}p_{[1][1]}   &  \cdots  &  p_{[1][0]}p_{[1][c-1]}   \\
 p_{[1][1]}p_{[1][0]}     &   p_{[1][1]}(p_{[1][1]} - 1)     &  \cdots  &  p_{[1][1]}p_{[1][c-1]}   \\
 \vdots    &   \vdots   &  \ddots  & \vdots   \\
 p_{[1][c-1]}p_{[1][0]}     &    p_{[1][c-1]}p_{[1][1]}    &  \cdots  & p_{[1][c-1]}(p_{[1][c-1]} - 1)   \\
 \end{bmatrix}^{-1}
 \otimes
   [ {\textsl{x}_1}^T\textsl{x}_1 ]^{-1}.
\end{align*} 

B\"ohning ~\citep{bohning1992multinomial} presented another  lemma about kronecker product: If  $A \le B$ in the Loewner order then  \begin{align} A \otimes C \le  B \otimes C  \label{Bohning inequality property }\end{align} for any symmetric, nonnegative definite $C$.

To apply the quadratic gradient in the multiclass LR, we need to build the matrix $\bar B$ first, trying to find a fixed Hessian matrix. We could try to build it directly from the Hessian matrix $\nabla^2$ based on the construction method of quadratic gradient:
\begin{align*}
 \bar B &=  
  \begin{bmatrix}
 \bar B_{[0]} & \mathbf{0} & \cdots & \mathbf 0   \\
 \mathbf 0        & \bar B_{[1]} & \cdots & \mathbf 0   \\
 \vdots  & \vdots & \ddots & \vdots       \\
 \mathbf 0 & \mathbf 0 & \cdots & \bar B_{[c-1]}   \\
 \end{bmatrix} 
 \Big (  \bar B_{[i]} =  
  \begin{bmatrix}
 \bar B_{[i][0]} & 0 & \cdots &  0   \\
 0        & \bar B_{[i][1]} & \cdots & 0   \\
 \vdots  & \vdots & \ddots & \vdots       \\
  0 & 0 & \cdots & \bar B_{[i][d]}   \\
 \end{bmatrix}  \Big )  
  \\
 &=
diag\{\bar B_{[0][0]},\bar B_{[0][1]},\cdots,\bar B_{[0][d]},\bar B_{[1][0]},\bar B_{[1][1]},\cdots,\bar B_{[1][d]},\cdots,\bar B_{[c-1][0]},\bar B_{[c-1][1]},\cdots,\bar B_{[c-1][d]} \} ,
\end{align*} 
where
$ \bar B_{[i][j]} =
 [|p_{[1][j]}p_{[1][0]}| +   |p_{[1][j]}p_{[1][1]}|  +  \cdots + |p_{[1][j]}(p_{[1][j]} - 1)| +  |p_{[1][j]}p_{[1][j+1]}| +\cdots + |p_{[1][j]}p_{[1][c-1]}| ] \times (|x_{[1][j]}x_{[1][0]}|+\hdots+|x_{[1][j]}x_{[1][d]}|)   + \epsilon .$

Since $0 < p_{[i][j]} < 1$ and  $\sum_{j=0}^{c-1} p_{[i][j]} = Prob(y=0|\textsl x= \textsl  x_{i}) + \dots + Prob(y=c-1|\textsl x= \textsl x_{i}) = 1 \ \text{for}\  1 \le i \le n $,  we get: 

\begin{equation*}
  \begin{aligned}
\bar B_{[i][j]}
 &=
 [|p_{[1][j]}p_{[1][0]}| +   |p_{[1][j]}p_{[1][1]}|  +  \cdots + |p_{[1][j]}(p_{[1][j]} - 1)| \\
 & \quad +  |p_{[1][j]}p_{[1][j+1]}| +\cdots + |p_{[1][j]}p_{[1][c-1]}| ] \times (|x_{[1][j]}x_{[1][0]}|+\hdots+|x_{[1][j]}x_{[1][d]}|)    + \epsilon  \\
 &=
 [p_{[1][j]}p_{[1][0]} +   p_{[1][j]}p_{[1][1]}  +  \cdots + p_{[1][j]}(1 - p_{[1][j]} ) \\
 & \quad +  p_{[1][j]}p_{[1][j+1]} +\cdots + p_{[1][j]}p_{[1][c-1]} ] \times (|x_{[1][j]}x_{[1][0]}|+\hdots+|x_{[1][j]}x_{[1][d]}|)   + \epsilon  \\ 
 &=
 [p_{[1][j]}(1 - p_{[1][j]} + 1 - p_{[1][j]} )] \times (|x_{[1][j]}x_{[1][0]}|+\hdots+|x_{[1][j]}x_{[1][d]}|)    + \epsilon \\
 &=
 2 \cdot [p_{[1][j]}(1 - p_{[1][j]} )] \times (|x_{[1][j]}x_{[1][0]}|+\hdots+|x_{[1][j]}x_{[1][d]}|)   + \epsilon \\
 &\le 
 0.5 \times (|x_{[1][j]}x_{[1][0]}|+\hdots+|x_{[1][j]}x_{[1][d]}|) + \epsilon  . \\
 \end{aligned}
\end{equation*}

The dataset $X$ should be normalized into the range $[0,1]$ in advance, 
further leading $\bar B_{[i][j]}$ to:
$\bar B_{[i][j]} \le 
 0.5 \times (x_{[0][0]}x_{[0][0]}+\hdots+x_{[0][0]}x_{[0][d]})   + \epsilon . $
It is straightforward to verify that setting $\bar B_{[i][j]} = 
 0.5 \times (x_{[0][0]}x_{[0][0]}+\hdots+x_{[0][0]}x_{[0][d]})  + \epsilon $ also meets the convergence condition of fixed Hessian Newton's method , in which case $\bar B$ could be built directly from $\bar H = \frac{1}{2}E \otimes \textsl x_1^T \textsl x_1$.

For the common cases where the dataset $X$ is a sample of size $n$:
\begin{align*}
 \nabla &= \sum_{i=1}^{n} [\bar y_{[i]} - p_{[i]}]^T \cdot \mathbf{x}_{[i]} = [\bar Y - P]^T \cdot X, \\
 \frac{ \partial^2 \ln L }{  \partial \pi \partial \pi }  &= \sum_{i=1}^{n} 
   \begin{bmatrix}
 p_{[i][0]}(p_{[i][0]} - 1)   &    p_{[i][0]}p_{[i][1]}   &  \cdots  &  p_{[i][0]}p_{[i][c-1]}   \\
 p_{[i][1]}p_{[i][0]}     &   p_{[i][1]}(p_{[i][1]} - 1)     &  \cdots  &  p_{[i][1]}p_{[i][c-1]}   \\
 \vdots    &   \vdots   &  \ddots  & \vdots   \\
 p_{[i][c-1]}p_{[i][0]}     &    p_{[i][c-1]}p_{[i][1]}    &  \cdots  & p_{[i][c-1]}(p_{[i][c-1]} - 1)   \\
 \end{bmatrix}
 \otimes
   [ {\textsl{x}_i}^T\textsl{x}_i ]  \\
 &= \Big [ P^TP - diag\{\sum_{i=1}^{n} p_{[i][0]},  \sum_{i=1}^{n} p_{[i][1]},  \cdots,  \sum_{i=1}^{n} p_{[i][d]}  \}  \Big ] \otimes [X^TX].
\end{align*} 

With the lemma $(\ref{Bohning inequality property })$ and  other kronecker product properties we could conclude:

\begin{align*}
 \bar B_{[i][j]}  &\le \sum_{i=1}^{n} 
  0.5 \times (x_{[i][0]}x_{[i][0]}+\hdots+x_{[i][0]}x_{[i][d]})  + \epsilon  \\
  &\le
   0.5 \times \sum_{i=1}^{n}(x_{[i][0]}x_{[i][0]}+\hdots+x_{[i][0]}x_{[i][d]})  + \epsilon.
\end{align*} 
That is, $\bar B$ can also be built from $\bar H = \frac{1}{2}E \otimes X^TX$
 and  B\"ohning and Lindsay \cite{bohning1988monotonicity} gave the same matrix as a good lower bound for $\textsl{ cox proportional hazards model }$. We thus obtain the quadratic gradicent $G = \bar B^{-1} \times \nabla$ and the iteration formula of the naive quadratic gradient ascent method in vector form: $\pi^T = \pi^T + G$ or in matrix form: $W = W + MG$ where $MG_{[i][j]} = \Bar B_{[i(1+d)+j][i(1+d)+j]}$.

Algorithm~\ref{ alg:enhanced NAG's algorithm } and Algorithm~\ref{ alg:enhanced Adagrad's algorithm } describe the enhanced NAG method and the enhanced Adagrad algorithm via quadratic gradient respectively.

\begin{algorithm}[ht]
    \caption{The enhanced Nesterov's accelerated gradient method  for Multiclass LR Training}
     \begin{algorithmic}[1]
        \Require training dataset $ X \in \mathbb{R} ^{n \times (1+d)} $; one-hot encoding training label $ Y \in \mathbb{R} ^{n \times c} $; and the number  $\kappa$ of iterations;
        \Ensure the parameter matrix $ V \in \mathbb{R} ^{c \times (1+d)} $ of the multiclass LR 
        
        \State Set $\bar H \gets -\frac{1}{2}X^TX$ 
        \Comment{$\bar H \in \mathbb R^{(1+d) \times (1+d)}$}

        \State Set $ V \gets \boldsymbol 0$, $ W \gets \boldsymbol 0$, $\bar B \gets  \boldsymbol 0$
        \Comment{$V \in \mathbb{R} ^{c \times (1+d)} $, $W \in \mathbb{R} ^{c \times (1+d)} $, $\bar B \in \mathbb R^{c \times (1+d)}$}
           \For{$j := 0$ to $d$}
              \State $\bar B[0][j] \gets \epsilon$
              \Comment{$\epsilon$ is a small positive constant such as $1e-8$}
              \For{$i := 0$ to $d$}
                 \State $ \bar B[0][j] \gets \bar B[0][j] + |\bar H[i][j]| $
              \EndFor
              \For{$i := 1$ to $c-1$}
                 \State $ \bar B[i][j] \gets \bar B[0][j]  $

              \EndFor
              \For{$i := 0$ to $c-1$}
                 \State $ \bar B[i][j] \gets 1.0 / \bar B[i][j]  $

              \EndFor
           \EndFor
       
        \State Set $\alpha_0 \gets 0.01$, $\alpha_1 \gets 0.5 \times (1 + \sqrt{1 + 4 \times \alpha_0^2} )$

        \For{$count := 1$ to $\kappa$}
           \State Set $Z \gets X \times V^T $
           \Comment{$Z \in \mathbb{R}^{n \times c}$  and $V^T$ means the transpose of matrix V}
           \For{$i := 1$ to $n$}
           \Comment{ $Z$ is going to store the inputs to the softmax function }
              \State Set $rowsum \gets 0$
              \For{$j := 0$ to $d$}
                 \State $ Z[i][j] \gets  e^{Z[i][j]} $ 
                 \Comment{ $e^{Z[i][j]}$ is to compute $\exp \{Z[i][j]\}$ } 
                 \State $rowsum \gets rowsum + Z[i][j]$ 
              \EndFor
              \For{$j := 0$ to $d$}
                 \State $ Z[i][j] \gets  Z[i][j] / rowsum $
                 \Comment{ $Z$ now  stores the outputs of the softmax function }
              \EndFor
           \EndFor
           \State Set $\boldsymbol g \gets (Y - Z)^T \times X$
           \Comment{ $\boldsymbol g \in \mathbb{R} ^{c \times (1+d)}$ }
           \State Set $ G \gets \boldsymbol 0$
           \For{$i := 0$ to $c-1$}
           	  \For{$j := 0$ to $d$}
                 \State $ G[i][j] \gets \bar B[i][j] \times \boldsymbol g[i][j]$
              \EndFor
           \EndFor
           \State Set $\eta \gets (1 - \alpha_0) / \alpha_1$, $\gamma \gets  1 / ( n \times count )$
           \Comment{$n$ is the size of training data}
           
              \State $ w_{temp} \gets W + (1 + \gamma)  \times G $
              \State $ W \gets (1 - \eta) \times w_{temp} + \eta \times V $
              \State $ V \gets w_{temp} $

	       \State $\alpha_0 \gets \alpha_1$, $\alpha_1 \gets 0.5 \times (1 + \sqrt{1 + 4 \times \alpha_0^2} )$ 
        \EndFor
        \State \Return $ W $
        \end{algorithmic}
       \label{ alg:enhanced NAG's algorithm }
\end{algorithm}

\begin{algorithm}[ht]
    \caption{The enhanced Adaptive Gradient Algorithm  for Multiclass LR Training}
     \begin{algorithmic}[1]
        \Require training dataset $ X \in \mathbb{R} ^{n \times (1+d)} $; one-hot encoding training label $ Y \in \mathbb{R} ^{n \times c} $; and the number  $\kappa$ of iterations;
        \Ensure the parameter matrix $ V \in \mathbb{R} ^{c \times (1+d)} $ of the multiclass LR 
        
        \State Set $\bar H \gets -\frac{1}{2}X^TX$ 
        \Comment{$\bar H \in \mathbb R^{(1+d) \times (1+d)}$}

        \State Set $ W \gets \boldsymbol 0$, $\bar B \gets  \boldsymbol 0$
        \Comment{$W \in \mathbb{R} ^{c \times (1+d)} $, $\bar B \in \mathbb R^{c \times (1+d)}$}
           \For{$j := 0$ to $d$}
              \State $\bar B[0][j] \gets \epsilon$
              \Comment{$\epsilon$ is a small positive constant such as $1e-8$}
              \For{$i := 0$ to $d$}
                 \State $ \bar B[0][j] \gets \bar B[0][j] + |\bar H[i][j]| $
              \EndFor
              \For{$i := 1$ to $c-1$}
                 \State $ \bar B[i][j] \gets \bar B[0][j]  $

              \EndFor
              \For{$i := 0$ to $c-1$}
                 \State $ \bar B[i][j] \gets 1.0 / \bar B[i][j]  $

              \EndFor
           \EndFor

        \State Set  $ Gt \gets \boldsymbol 0$
        \Comment{$Gt \in \mathbb{R} ^{c \times (1+d)} $}
        \State Set $\alpha_0 \gets 0.01$, $\alpha_1 \gets 0.5 \times (1 + \sqrt{1 + 4 \times \alpha_0^2} )$

        \For{$count := 1$ to $\kappa$}
           \State Set $Z \gets X \times W^T $
           \Comment{$Z \in \mathbb{R}^{n \times c}$  and $W^T$ means the transpose of matrix W}
           \For{$i := 1$ to $n$}
           \Comment{ $Z$ is going to store the inputs to the softmax function }
              \State Set $rowsum \gets 0$
              \For{$j := 0$ to $d$}
                 \State $ Z[i][j] \gets  e^{Z[i][j]} $ 
                 \Comment{ $e^{Z[i][j]}$ is to compute $\exp \{Z[i][j]\}$ } 
                 \State $rowsum \gets rowsum + Z[i][j]$ 
              \EndFor
              \For{$j := 0$ to $d$}
                 \State $ Z[i][j] \gets  Z[i][j] / rowsum $
                 \Comment{ $Z$ now  stores the outputs of the softmax function }
              \EndFor
           \EndFor
           \State Set $\boldsymbol g \gets (Y - Z)^T \times X$
           \Comment{ $\boldsymbol g \in \mathbb{R} ^{c \times (1+d)}$ }
           \State Set $ G \gets \boldsymbol 0$
           \For{$i := 0$ to $c-1$}
           	  \For{$j := 0$ to $d$}
                 \State $ G[i][j] \gets \bar B[i][j] \times \boldsymbol g[i][j]$
              \EndFor
           \EndFor
           \For{$i := 0$ to $c-1$}
           	  \For{$j := 0$ to $d$}
                 \State $ Gt[i][j] \gets Gt[i][j] + G[i][j] \times G[i][j]$
              \EndFor
           \EndFor
           \State Set  $ Gamma \gets \boldsymbol 0$
           \Comment{$Gamma \in \mathbb{R} ^{c \times (1+d)} $}
           \For{$i := 0$ to $c-1$}
           	  \For{$j := 0$ to $d$}
                 \State $ Gamma[i][j] \gets (1.0 +0.01) /  \sqrt{ \epsilon + Gt[i][j] } $
              \EndFor
           \EndFor
           \LineComment{To update the weight vector $W$ }
           \For{$i := 0$ to $c-1$}
           	  \For{$j := 0$ to $d$}
                 \State $ W[i][j] \gets W[i][j] + Gamma[i][j] \times G[i][j] $
              \EndFor
           \EndFor
        \EndFor
        \State \Return $ W $
        \end{algorithmic}
       \label{ alg:enhanced Adagrad's algorithm }
\end{algorithm}

\section{Experiments}
All the python source code to implement the experiments in the paper  is openly available at: \href{https://github.com/petitioner/ML.MulticlassLRtraining}{$\texttt{https://github.com/petitioner/ML.MulticlassLRtraining}$}  . 

To study the performance of the enhanced NAG and Adagrad methods for multiclass LR,  we consider three datasets adopted by~\cite{abramovich2021multiclass}: vehicle, shuttle, and segmentation taken from LIBSVM Data\footnote{ \href{https://www.csie.ntu.edu.tw/~cjlin/libsvmtools/datasets/multiclass.html}{$\texttt{https://www.csie.ntu.edu.tw/~cjlin/libsvmtools/datasets/multiclass.html}$} }.
Table \ref{tab2} describes the three datasets. We also used the testing data of the   shuttle dataset in the training stage.
Maximum likelihood estimation (the loss function) and accuracy are selected as the only indicators. 
We compare  the performance of the enhanced methods with their corresponding first-order naive gradient method. 
The accuracy and maximum likelihood estimation are reported in Figures \ref{fig0} and \ref{fig1}. We can remark that the performances of the enhanced methods outperform their original ones.  



\begin{table}[htbp]
\centering
\caption{Characteristics of the several datasets used in our experiments}
\label{tab2}
\begin{tabular}{|c|c|c|c|c|c|c|c|c|}
\hline
Dataset &   \mysplit{No. Samples  \\ (training) }   &   \mysplit{ No. Samples  \\ (testing)  }   &  No. Features   & No. Classes  \\
\hline
iris &   150   &   N/A   &  4   & 3  \\
\hline
segment &   2, 310   &   N/A   &  19   & 7  \\
\hline
shuttle &   43, 500   &   14, 500   &  9   & 7  \\
\hline
vehicle &   846   &   N/A   &  18   & 4  \\
\hline
\end{tabular}
\end{table}

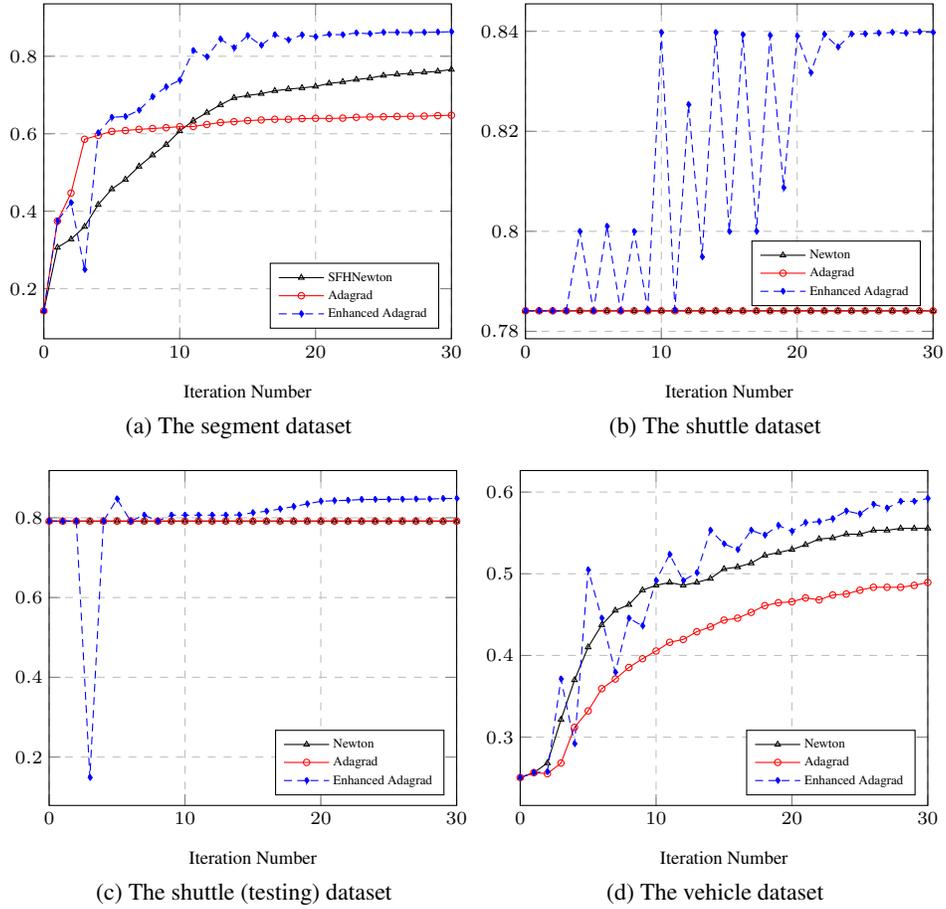
\begin{figure}[ht]
\centering
\captionsetup[subfigure]{justification=centering}
\subfloat[The segment dataset]{%
\begin{tikzpicture}
\scriptsize  
\begin{axis}[
    width=7cm,
    xlabel={Iteration Number},
    xmin=0, xmax=30,
    legend pos=south east,
    legend style={nodes={scale=0.7, transform shape}},
    legend cell align={left},
    xmajorgrids=true,
    ymajorgrids=true,
    grid style=dashed,
]
\addplot[
    color=black,
    mark=triangle,
    mark size=1.2pt,
    ] 
    table [x=Iterations, y=SFHNewton, col sep=comma] {PythonExperiment_SFHNewtonvs.Adagradvs.EnAdagrad_PREC_training_segment.scale.csv};
\addplot[
    color=red,
    mark=o,
    mark size=1.2pt,
    ] 
    table [x=Iterations, y=Adagrad, col sep=comma] {PythonExperiment_SFHNewtonvs.Adagradvs.EnAdagrad_PREC_training_segment.scale.csv};
\addplot[
    color=blue,
    mark=diamond*, 
    mark size=1.2pt,
    densely dashed
    ]  
    table [x=Iterations, y=AdagradQG, col sep=comma] {PythonExperiment_SFHNewtonvs.Adagradvs.EnAdagrad_PREC_training_segment.scale.csv};
   \addlegendentry{SFHNewton}   
   \addlegendentry{Adagrad}
   \addlegendentry{Enhanced Adagrad}
\end{axis}
\end{tikzpicture}
\label{fig:subfig01}}
\subfloat[The shuttle dataset]{%
\begin{tikzpicture}
\scriptsize  
\begin{axis}[
    width=7cm,
    xlabel={Iteration Number},
    xmin=0, xmax=30,
    legend pos=south east,
    legend style={yshift=0.3cm, nodes={scale=0.7, transform shape}},
    legend cell align={left},
    xmajorgrids=true,
    ymajorgrids=true,
    grid style=dashed,
]
\addplot[
    color=black,
    mark=triangle,
    mark size=1.2pt,
    ] 
    table [x=Iterations, y=SFHNewton, col sep=comma] {PythonExperiment_SFHNewtonvs.Adagradvs.EnAdagrad_PREC_training_shuttle.scale.csv};
\addplot[
    color=red,
    mark=o,
    mark size=1.2pt,
    ] 
    table [x=Iterations, y=Adagrad, col sep=comma] {PythonExperiment_SFHNewtonvs.Adagradvs.EnAdagrad_PREC_training_shuttle.scale.csv};
\addplot[
    color=blue,
    mark=diamond*, 
    mark size=1.2pt,
    densely dashed
    ]  
    table [x=Iterations, y=AdagradQG, col sep=comma] {PythonExperiment_SFHNewtonvs.Adagradvs.EnAdagrad_PREC_training_shuttle.scale.csv};
   \addlegendentry{Newton}   
   \addlegendentry{Adagrad}
   \addlegendentry{Enhanced Adagrad}
\end{axis}
\end{tikzpicture}
\label{fig:subfig02}}

\subfloat[The shuttle (testing) dataset]{%
\begin{tikzpicture}
\scriptsize  
\begin{axis}[
    width=7cm,
    xlabel={Iteration Number},
    xmin=0, xmax=30,
    legend pos=south east,
    legend style={nodes={scale=0.7, transform shape}},
    legend cell align={left},
    xmajorgrids=true,
    ymajorgrids=true,
    grid style=dashed,
]
\addplot[
    color=black,
    mark=triangle,
    mark size=1.2pt,
    ] 
    table [x=Iterations, y=SFHNewton, col sep=comma] {PythonExperiment_SFHNewtonvs.Adagradvs.EnAdagrad_PREC_training_shuttle.scale.t.csv};
\addplot[
    color=red,
    mark=o,
    mark size=1.2pt,
    ] 
    table [x=Iterations, y=Adagrad, col sep=comma] {PythonExperiment_SFHNewtonvs.Adagradvs.EnAdagrad_PREC_training_shuttle.scale.t.csv};
\addplot[
    color=blue,
    mark=diamond*, 
    mark size=1.2pt,
    densely dashed
    ]  
    table [x=Iterations, y=AdagradQG, col sep=comma] {PythonExperiment_SFHNewtonvs.Adagradvs.EnAdagrad_PREC_training_shuttle.scale.t.csv};
   \addlegendentry{Newton}   
   \addlegendentry{Adagrad}
   \addlegendentry{Enhanced Adagrad}
\end{axis}
\end{tikzpicture}
\label{fig:subfig03}}
\subfloat[The vehicle dataset]{%
\begin{tikzpicture}
\scriptsize  
\begin{axis}[
    width=7cm,
    xlabel={Iteration Number},
    xmin=0, xmax=30,
    legend pos=south east,
    legend style={nodes={scale=0.7, transform shape}},
    legend cell align={left},
    xmajorgrids=true,
    ymajorgrids=true,
    grid style=dashed,
]
\addplot[
    color=black,
    mark=triangle,
    mark size=1.2pt,
    ] 
    table [x=Iterations, y=SFHNewton, col sep=comma] {PythonExperiment_SFHNewtonvs.Adagradvs.EnAdagrad_PREC_training_vehicle.scale.csv};
\addplot[
    color=red,
    mark=o,
    mark size=1.2pt,
    ] 
    table [x=Iterations, y=Adagrad, col sep=comma] {PythonExperiment_SFHNewtonvs.Adagradvs.EnAdagrad_PREC_training_vehicle.scale.csv};
\addplot[
    color=blue,
    mark=diamond*, 
    mark size=1.2pt,
    densely dashed
    ]  
    table [x=Iterations, y=AdagradQG, col sep=comma] {PythonExperiment_SFHNewtonvs.Adagradvs.EnAdagrad_PREC_training_vehicle.scale.csv};
   \addlegendentry{Newton}   
   \addlegendentry{Adagrad}
   \addlegendentry{Enhanced Adagrad}
\end{axis}
\end{tikzpicture}
\label{fig:subfig04}}
\caption{\protect\centering Training results PREC for SFHNewton vs. Adagrad vs. Enhanced Adagrad}
\label{fig0}
\end{figure}

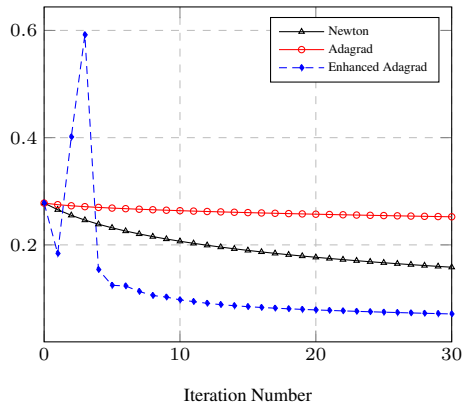
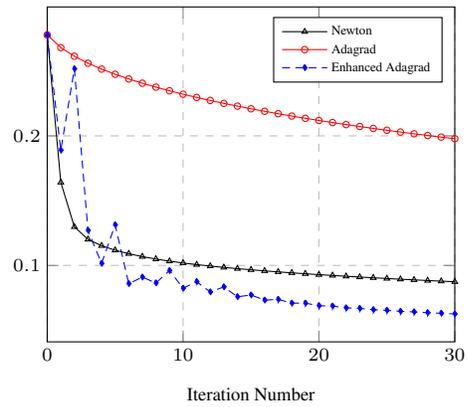
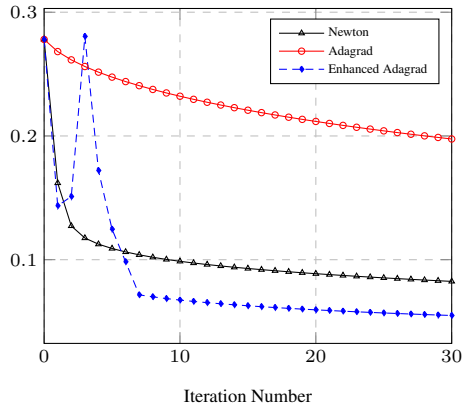
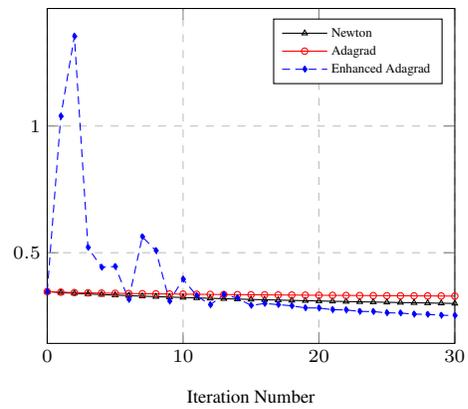
\begin{figure}[ht]
\centering
\captionsetup[subfigure]{justification=centering}
\subfloat[The segment dataset]{%
\begin{tikzpicture}
\scriptsize  
\begin{axis}[
    width=7cm,
    xlabel={Iteration Number},
    xmin=0, xmax=30,
    legend pos=north east,
    legend style={yshift=0cm,nodes={scale=0.7, transform shape}},
    legend cell align={left},
    xmajorgrids=true,
    ymajorgrids=true,
    grid style=dashed,
]
\addplot[
    color=black,
    mark=triangle,
    mark size=1.2pt,
    ] 
    table [x=Iterations, y=SFHNewton, col sep=comma] {PythonExperiment_SFHNewtonvs.Adagradvs.EnAdagrad_LOSS_training_segment.scale.csv};
\addplot[
    color=red,
    mark=o,
    mark size=1.2pt,
    ] 
    table [x=Iterations, y=Adagrad, col sep=comma] {PythonExperiment_SFHNewtonvs.Adagradvs.EnAdagrad_LOSS_training_segment.scale.csv};
\addplot[
    color=blue,
    mark=diamond*, 
    mark size=1.2pt,
    densely dashed
    ]  
    table [x=Iterations, y=AdagradQG, col sep=comma] {PythonExperiment_SFHNewtonvs.Adagradvs.EnAdagrad_LOSS_training_segment.scale.csv};
   \addlegendentry{Newton}   
   \addlegendentry{Adagrad}
   \addlegendentry{Enhanced Adagrad}
\end{axis}
\end{tikzpicture}
\label{fig:subfig11}}
\subfloat[The shuttle dataset]{%
\begin{tikzpicture}
\scriptsize  
\begin{axis}[
    width=7cm,
    xlabel={Iteration Number},
    xmin=0, xmax=30,
    legend pos=north east,
    legend style={yshift=0cm,nodes={scale=0.7, transform shape}},
    legend cell align={left},
    xmajorgrids=true,
    ymajorgrids=true,
    grid style=dashed,
]
\addplot[
    color=black,
    mark=triangle,
    mark size=1.2pt,
    ] 
    table [x=Iterations, y=SFHNewton, col sep=comma] {PythonExperiment_SFHNewtonvs.Adagradvs.EnAdagrad_LOSS_training_shuttle.scale.csv};
\addplot[
    color=red,
    mark=o,
    mark size=1.2pt,
    ] 
    table [x=Iterations, y=Adagrad, col sep=comma] {PythonExperiment_SFHNewtonvs.Adagradvs.EnAdagrad_LOSS_training_shuttle.scale.csv};
\addplot[
    color=blue,
    mark=diamond*, 
    mark size=1.2pt,
    densely dashed
    ]  
    table [x=Iterations, y=AdagradQG, col sep=comma] {PythonExperiment_SFHNewtonvs.Adagradvs.EnAdagrad_LOSS_training_shuttle.scale.csv};
   \addlegendentry{Newton}   
   \addlegendentry{Adagrad}
   \addlegendentry{Enhanced Adagrad}
\end{axis}
\end{tikzpicture}
\label{fig:subfig12}}

\subfloat[The shuttle (testing) dataset]{%
\begin{tikzpicture}
\scriptsize  
\begin{axis}[
    width=7cm,
    xlabel={Iteration Number},
    xmin=0, xmax=30,
    legend pos=north east,
    legend style={yshift=0cm,nodes={scale=0.7, transform shape}},
    legend cell align={left},
    xmajorgrids=true,
    ymajorgrids=true,
    grid style=dashed,
]
\addplot[
    color=black,
    mark=triangle,
    mark size=1.2pt,
    ] 
    table [x=Iterations, y=SFHNewton, col sep=comma] {PythonExperiment_SFHNewtonvs.Adagradvs.EnAdagrad_LOSS_training_shuttle.scale.t.csv};
\addplot[
    color=red,
    mark=o,
    mark size=1.2pt,
    ] 
    table [x=Iterations, y=Adagrad, col sep=comma] {PythonExperiment_SFHNewtonvs.Adagradvs.EnAdagrad_LOSS_training_shuttle.scale.t.csv};
\addplot[
    color=blue,
    mark=diamond*, 
    mark size=1.2pt,
    densely dashed
    ]  
    table [x=Iterations, y=AdagradQG, col sep=comma] {PythonExperiment_SFHNewtonvs.Adagradvs.EnAdagrad_LOSS_training_shuttle.scale.t.csv};
   \addlegendentry{Newton}   
   \addlegendentry{Adagrad}
   \addlegendentry{Enhanced Adagrad}
\end{axis}
\end{tikzpicture}
\label{fig:subfig13}}
\subfloat[The vehicle dataset]{%
\begin{tikzpicture}
\scriptsize  
\begin{axis}[
    width=7cm,
    xlabel={Iteration Number},
    xmin=0, xmax=30,
    legend pos=north east,
    legend style={yshift=0cm,nodes={scale=0.7, transform shape}},
    legend cell align={left},
    xmajorgrids=true,
    ymajorgrids=true,
    grid style=dashed,
]
\addplot[
    color=black,
    mark=triangle,
    mark size=1.2pt,
    ] 
    table [x=Iterations, y=SFHNewton, col sep=comma] {PythonExperiment_SFHNewtonvs.Adagradvs.EnAdagrad_LOSS_training_vehicle.scale.csv};
\addplot[
    color=red,
    mark=o,
    mark size=1.2pt,
    ] 
    table [x=Iterations, y=Adagrad, col sep=comma] {PythonExperiment_SFHNewtonvs.Adagradvs.EnAdagrad_LOSS_training_vehicle.scale.csv};
\addplot[
    color=blue,
    mark=diamond*, 
    mark size=1.2pt,
    densely dashed
    ]  
    table [x=Iterations, y=AdagradQG, col sep=comma] {PythonExperiment_SFHNewtonvs.Adagradvs.EnAdagrad_LOSS_training_vehicle.scale.csv};
   \addlegendentry{Newton}   
   \addlegendentry{Adagrad}
   \addlegendentry{Enhanced Adagrad}
\end{axis}
\end{tikzpicture}
\label{fig:subfig14}}
\caption{\protect\centering Training results LOSS for SFHNewton vs. Adagrad vs. Enhanced Adagrad}
\label{fig1}
\end{figure}


\begin{figure}[ht]
\centering
\captionsetup[subfigure]{justification=centering}
\subfloat[The segment dataset]{%
\begin{tikzpicture}
\scriptsize  
\begin{axis}[
    width=7cm,
    xlabel={Iteration Number},
    xmin=0, xmax=30,
    legend pos=south east,
    legend style={nodes={scale=0.7, transform shape}},
    legend cell align={left},
    xmajorgrids=true,
    ymajorgrids=true,
    grid style=dashed,
]
\addplot[
    color=black,
    mark=triangle,
    mark size=1.2pt,
    ] 
    table [x=Iterations, y=SFHN, col sep=comma] {PythonExperiment_SFHNvs.NAGvs.NAGQG_PREC_training_segment.scale.csv};
\addplot[
    color=red,
    mark=o,
    mark size=1.2pt,
    ] 
    table [x=Iterations, y=NAG, col sep=comma] {PythonExperiment_SFHNvs.NAGvs.NAGQG_PREC_training_segment.scale.csv};
\addplot[
    color=blue,
    mark=diamond*, 
    mark size=1.2pt,
    densely dashed
    ]  
    table [x=Iterations, y=NAGQG, col sep=comma] {PythonExperiment_SFHNvs.NAGvs.NAGQG_PREC_training_segment.scale.csv};
   \addlegendentry{SFH Newton}   
   \addlegendentry{NAG}
   \addlegendentry{Enhanced NAG}
\end{axis}
\end{tikzpicture}
\label{fig:subfig01}}
\subfloat[The shuttle dataset]{%
\begin{tikzpicture}
\scriptsize  
\begin{axis}[
    width=7cm,
    xlabel={Iteration Number},
    xmin=0, xmax=30,
    legend pos=south east,
    legend style={xshift=-.8cm,,yshift=+.5cm, nodes={scale=0.7, transform shape}},
    legend cell align={left},
    xmajorgrids=true,
    ymajorgrids=true,
    grid style=dashed,
]
\addplot[
    color=black,
    mark=triangle,
    mark size=1.2pt,
    ] 
    table [x=Iterations, y=SFHN, col sep=comma] {PythonExperiment_SFHNvs.NAGvs.NAGQG_PREC_training_shuttle.scale.csv};
\addplot[
    color=red,
    mark=o,
    mark size=1.2pt,
    ] 
    table [x=Iterations, y=NAG, col sep=comma] {PythonExperiment_SFHNvs.NAGvs.NAGQG_PREC_training_shuttle.scale.csv};
\addplot[
    color=blue,
    mark=diamond*, 
    mark size=1.2pt,
    densely dashed
    ]  
    table [x=Iterations, y=NAGQG, col sep=comma] {PythonExperiment_SFHNvs.NAGvs.NAGQG_PREC_training_shuttle.scale.csv};
   \addlegendentry{SFH Newton}   
   \addlegendentry{NAG}
   \addlegendentry{Enhanced NAG}
\end{axis}
\end{tikzpicture}
\label{fig:subfig02}}

\subfloat[The shuttle (testing) dataset]{%
\begin{tikzpicture}
\scriptsize  
\begin{axis}[
    width=7cm,
    xlabel={Iteration Number},
    xmin=0, xmax=30,
    legend pos=south east,
    legend style={xshift=-1.8cm,,yshift=+.5cm,nodes={scale=0.7, transform shape}},
    legend cell align={left},
    xmajorgrids=true,
    ymajorgrids=true,
    grid style=dashed,
]
\addplot[
    color=black,
    mark=triangle,
    mark size=1.2pt,
    ] 
    table [x=Iterations, y=SFHN, col sep=comma] {PythonExperiment_SFHNvs.NAGvs.NAGQG_PREC_training_shuttle.scale.t.csv};
\addplot[
    color=red,
    mark=o,
    mark size=1.2pt,
    ] 
    table [x=Iterations, y=NAG, col sep=comma] {PythonExperiment_SFHNvs.NAGvs.NAGQG_PREC_training_shuttle.scale.t.csv};
\addplot[
    color=blue,
    mark=diamond*, 
    mark size=1.2pt,
    densely dashed
    ]  
    table [x=Iterations, y=NAGQG, col sep=comma] {PythonExperiment_SFHNvs.NAGvs.NAGQG_PREC_training_shuttle.scale.t.csv};
   \addlegendentry{SFH Newton}   
   \addlegendentry{NAG}
   \addlegendentry{Enhanced NAG}
\end{axis}
\end{tikzpicture}
\label{fig:subfig03}}
\subfloat[The vehicle dataset]{%
\begin{tikzpicture}
\scriptsize  
\begin{axis}[
    width=7cm,
    xlabel={Iteration Number},
    xmin=0, xmax=30,
    legend pos=south east,
    legend style={nodes={scale=0.7, transform shape}},
    legend cell align={left},
    xmajorgrids=true,
    ymajorgrids=true,
    grid style=dashed,
]
\addplot[
    color=black,
    mark=triangle,
    mark size=1.2pt,
    ] 
    table [x=Iterations, y=SFHN, col sep=comma] {PythonExperiment_SFHNvs.NAGvs.NAGQG_PREC_training_vehicle.scale.csv};
\addplot[
    color=red,
    mark=o,
    mark size=1.2pt,
    ] 
    table [x=Iterations, y=NAG, col sep=comma] {PythonExperiment_SFHNvs.NAGvs.NAGQG_PREC_training_vehicle.scale.csv};
\addplot[
    color=blue,
    mark=diamond*, 
    mark size=1.2pt,
    densely dashed
    ]  
    table [x=Iterations, y=NAGQG, col sep=comma] {PythonExperiment_SFHNvs.NAGvs.NAGQG_PREC_training_vehicle.scale.csv};
   \addlegendentry{SFH Newton}   
   \addlegendentry{NAG}
   \addlegendentry{Enhanced NAG}
\end{axis}
\end{tikzpicture}
\label{fig:subfig04}}
\caption{\protect\centering Training results PREC for SFHNewton vs. Adagrad vs. Enhanced Adagrad}
\label{fig0}
\end{figure}
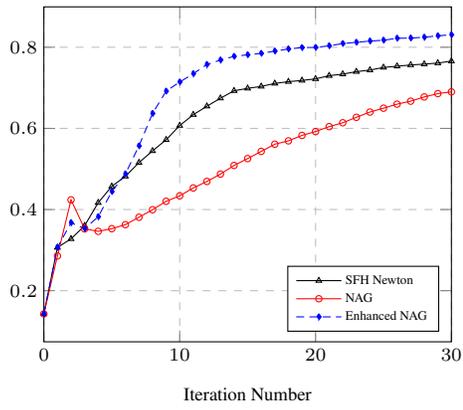
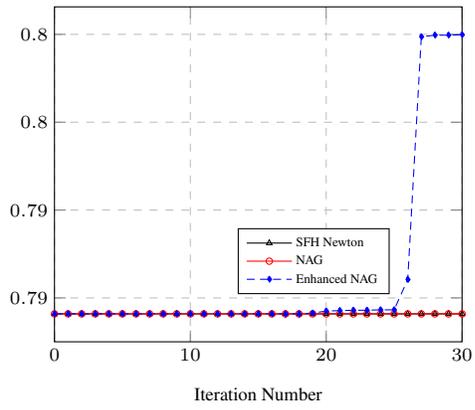
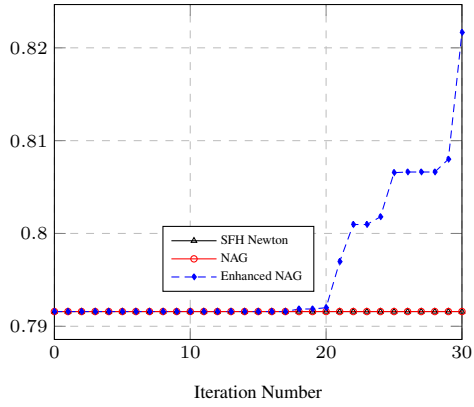
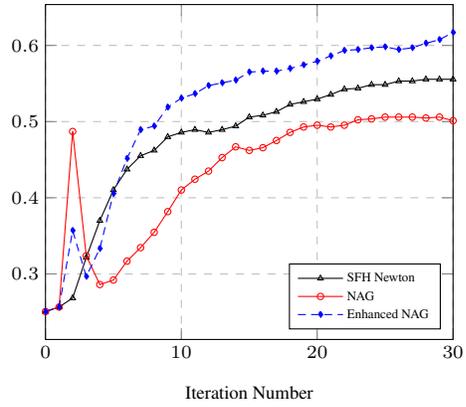

\begin{figure}[ht]
\centering
\captionsetup[subfigure]{justification=centering}
\subfloat[The segment dataset]{%
\begin{tikzpicture}
\scriptsize  
\begin{axis}[
    width=7cm,
    xlabel={Iteration Number},
    xmin=0, xmax=30,
    legend pos=north east,
    legend style={yshift=0cm,nodes={scale=0.7, transform shape}},
    legend cell align={left},
    xmajorgrids=true,
    ymajorgrids=true,
    grid style=dashed,
]
\addplot[
    color=black,
    mark=triangle,
    mark size=1.2pt,
    ] 
    table [x=Iterations, y=SFHN, col sep=comma] {PythonExperiment_SFHNvs.NAGvs.NAGQG_LOSS_training_segment.scale.csv};
\addplot[
    color=red,
    mark=o,
    mark size=1.2pt,
    ] 
    table [x=Iterations, y=NAG, col sep=comma] {PythonExperiment_SFHNvs.NAGvs.NAGQG_LOSS_training_segment.scale.csv};
\addplot[
    color=blue,
    mark=diamond*, 
    mark size=1.2pt,
    densely dashed
    ]  
    table [x=Iterations, y=NAGQG, col sep=comma] {PythonExperiment_SFHNvs.NAGvs.NAGQG_LOSS_training_segment.scale.csv};
   \addlegendentry{SFH Newton}   
   \addlegendentry{NAG}
   \addlegendentry{Enhanced NAG}
\end{axis}
\end{tikzpicture}
\label{fig:subfig11}}
\subfloat[The shuttle dataset]{%
\begin{tikzpicture}
\scriptsize  
\begin{axis}[
    width=7cm,
    xlabel={Iteration Number},
    xmin=0, xmax=30,
    legend pos=north east,
    legend style={yshift=0cm,nodes={scale=0.7, transform shape}},
    legend cell align={left},
    xmajorgrids=true,
    ymajorgrids=true,
    grid style=dashed,
]
\addplot[
    color=black,
    mark=triangle,
    mark size=1.2pt,
    ] 
    table [x=Iterations, y=SFHN, col sep=comma] {PythonExperiment_SFHNvs.NAGvs.NAGQG_LOSS_training_shuttle.scale.csv};
\addplot[
    color=red,
    mark=o,
    mark size=1.2pt,
    ] 
    table [x=Iterations, y=NAG, col sep=comma] {PythonExperiment_SFHNvs.NAGvs.NAGQG_LOSS_training_shuttle.scale.csv};
\addplot[
    color=blue,
    mark=diamond*, 
    mark size=1.2pt,
    densely dashed
    ]  
    table [x=Iterations, y=NAGQG, col sep=comma] {PythonExperiment_SFHNvs.NAGvs.NAGQG_LOSS_training_shuttle.scale.csv};
   \addlegendentry{SFH Newton}   
   \addlegendentry{NAG}
   \addlegendentry{Enhanced NAG}
\end{axis}
\end{tikzpicture}
\label{fig:subfig12}}

\subfloat[The shuttle (testing) dataset]{%
\begin{tikzpicture}
\scriptsize  
\begin{axis}[
    width=7cm,
    xlabel={Iteration Number},
    xmin=0, xmax=30,
    legend pos=north east,
    legend style={yshift=0cm,nodes={scale=0.7, transform shape}},
    legend cell align={left},
    xmajorgrids=true,
    ymajorgrids=true,
    grid style=dashed,
]
\addplot[
    color=black,
    mark=triangle,
    mark size=1.2pt,
    ] 
    table [x=Iterations, y=SFHN, col sep=comma] {PythonExperiment_SFHNvs.NAGvs.NAGQG_LOSS_training_shuttle.scale.t.csv};
\addplot[
    color=red,
    mark=o,
    mark size=1.2pt,
    ] 
    table [x=Iterations, y=NAG, col sep=comma] {PythonExperiment_SFHNvs.NAGvs.NAGQG_LOSS_training_shuttle.scale.t.csv};
\addplot[
    color=blue,
    mark=diamond*, 
    mark size=1.2pt,
    densely dashed
    ]  
    table [x=Iterations, y=NAGQG, col sep=comma] {PythonExperiment_SFHNvs.NAGvs.NAGQG_LOSS_training_shuttle.scale.t.csv};
   \addlegendentry{SFH Newton}   
   \addlegendentry{NAG}
   \addlegendentry{Enhanced NAG}
\end{axis}
\end{tikzpicture}
\label{fig:subfig13}}
\subfloat[The vehicle dataset]{%
\begin{tikzpicture}
\scriptsize  
\begin{axis}[
    width=7cm,
    xlabel={Iteration Number},
    xmin=0, xmax=30,
    legend pos=north east,
    legend style={yshift=0cm,nodes={scale=0.7, transform shape}},
    legend cell align={left},
    xmajorgrids=true,
    ymajorgrids=true,
    grid style=dashed,
]
\addplot[
    color=black,
    mark=triangle,
    mark size=1.2pt,
    ] 
    table [x=Iterations, y=SFHN, col sep=comma] {PythonExperiment_SFHNvs.NAGvs.NAGQG_LOSS_training_vehicle.scale.csv};
\addplot[
    color=red,
    mark=o,
    mark size=1.2pt,
    ] 
    table [x=Iterations, y=NAG, col sep=comma] {PythonExperiment_SFHNvs.NAGvs.NAGQG_LOSS_training_vehicle.scale.csv};
\addplot[
    color=blue,
    mark=diamond*, 
    mark size=1.2pt,
    densely dashed
    ]  
    table [x=Iterations, y=NAGQG, col sep=comma] {PythonExperiment_SFHNvs.NAGvs.NAGQG_LOSS_training_vehicle.scale.csv};
   \addlegendentry{SFH Newton}   
   \addlegendentry{NAG}
   \addlegendentry{Enhanced NAG}
\end{axis}
\end{tikzpicture}
\label{fig:subfig14}}
\caption{\protect\centering Training results LOSS for SFHNewton vs. NAG vs. NAGG}
\label{fig1}
\end{figure}
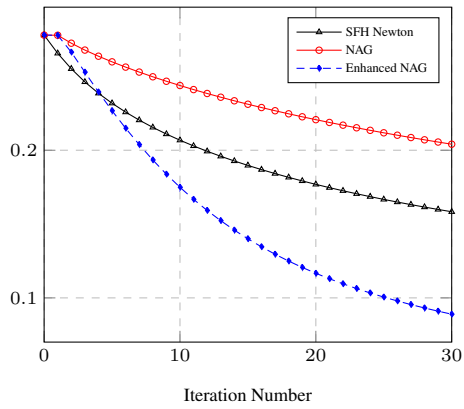
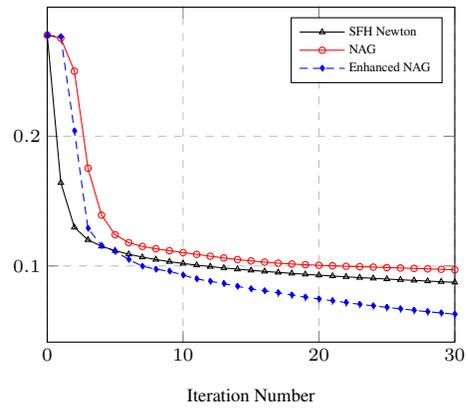
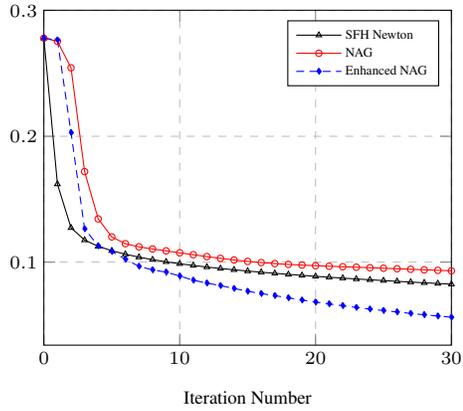
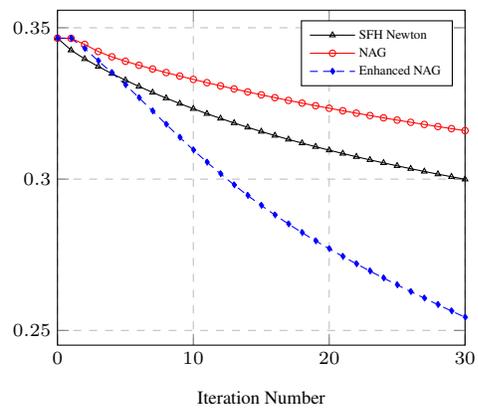

\section{Conclusion}


There is a good chance that the enhanced gradient methods for multiclass LR could be used in the classisation neural-network training via the softmax activation and the cross-entropy loss.  




\bibliography{ML.MulticlassLRtraining}

\begin{thebibliography}{}

\bibitem[Abramovich et~al., 2021]{abramovich2021multiclass}
Abramovich, F., Grinshtein, V., and Levy, T. (2021).
\newblock Multiclass classification by sparse multinomial logistic regression.
\newblock {\em IEEE Transactions on Information Theory}, 67(7):4637--4646.

\bibitem[B{\"o}hning, 1992]{bohning1992multinomial}
B{\"o}hning, D. (1992).
\newblock Multinomial logistic regression algorithm.
\newblock {\em Annals of the institute of Statistical Mathematics},
  44(1):197--200.

\bibitem[B{\"o}hning and Lindsay, 1988]{bohning1988monotonicity}
B{\"o}hning, D. and Lindsay, B.~G. (1988).
\newblock Monotonicity of quadratic-approximation algorithms.
\newblock {\em Annals of the Institute of Statistical Mathematics},
  40(4):641--663.

\bibitem[Bonte and Vercauteren, 2018]{bonte2018privacy}
Bonte, C. and Vercauteren, F. (2018).
\newblock Privacy-preserving logistic regression training.
\newblock {\em BMC medical genomics}, 11(4):13--21.

\bibitem[Chiang, 2022a]{chiang2022novel}
Chiang, J. (2022a).
\newblock A novel matrix-encoding method for privacy-preserving neural networks
  (inference).
\newblock {\em arXiv preprint arXiv:2201.12577}.

\bibitem[Chiang, 2022b]{chiang2022privacy}
Chiang, J. (2022b).
\newblock Privacy-preserving logistic regression training with a faster
  gradient variant.
\newblock {\em arXiv preprint arXiv:2201.10838}.

\bibitem[Smart and Vercauteren, 2011]{SmartandVercauteren_SIMD}
Smart, N. and Vercauteren, F. (2011).
\newblock Fully homomorphic simd operations.
\newblock Cryptology ePrint Archive, Report 2011/133.
\newblock \url{https://ia.cr/2011/133}.

\end{thebibliography}
\bibliographystyle{apalike}  

\end{document}